\documentclass{article}

\PassOptionsToPackage{numbers}{natbib}
\usepackage[utf8]{inputenc}  
\usepackage[T1]{fontenc}    
\usepackage{hyperref}       
\usepackage{url}            

\usepackage{breakurl}
\usepackage{booktabs}       
\usepackage{amsfonts}       
\usepackage{nicefrac}       
\usepackage{microtype}      
\usepackage{todonotes}

\usepackage[final]{apt_2022}

\usepackage{bm}
\usepackage{amsthm}
\usepackage{amsmath}
\usepackage{subcaption}
\usepackage{graphicx}

\usepackage{floatflt}
\usepackage{graphics}
\usepackage{siunitx}
\usepackage{wrapfig}

\usepackage{grffile}
\usepackage{arydshln}
\usepackage{siunitx}

\usepackage{multirow}

\title{TWIZ-v2: The Wizard of Multimodal Conversational-Stimulus}

\author{
Rafael Ferreira, Diogo Tavares, Diogo Silva, Rodrigo Valério, João Bordalo,
\\ \textbf{Inês Simões, Vasco Ramos, David Semedo, Joao Magalhaes}
\\ NOVA University of Lisbon, NOVA LINCS \\
\texttt{\{rah.ferreira,dc.tavares,dmgc.silva,r.valerio}\\
\texttt{j.bordalo,ir.simoes,vcc.ramos\}@campus.fct.unl.pt}
\\
\texttt{\{df.semedo,jm.magalhaes\}@fct.unl.pt}
}

\begin{document}

\maketitle

\begin{abstract}
In this report, we describe the vision, challenges, and scientific contributions of the Task Wizard team, TWIZ, in the Alexa Prize TaskBot Challenge 2022~\cite{taskbot_2023}.
Our vision, is to build TWIZ bot as an \textit{helpful}, \textit{multimodal}, \textit{knowledgeable}, and \textit{engaging} assistant that can guide users towards the successful completion of complex manual tasks.
To achieve this, we focus our efforts on three main research questions: (1) Humanly-Shaped Conversations, by providing information in a knowledgeable way; (2) Multimodal Stimulus, making use of various modalities including voice, images, and videos; and (3) Zero-shot Conversational Flows, to improve the robustness of the interaction to unseen scenarios.
TWIZ is an assistant capable of supporting a wide range of tasks, with several innovative features such as creative cooking, video navigation through voice, and the robust TWIZ-LLM, a Large Language Model trained for dialoguing about complex manual tasks.
Given ratings and feedback provided by users, we observed that TWIZ bot is an effective and robust system, capable of guiding users through tasks while providing several multimodal stimuli.
\end{abstract}

\section{Introduction}
\label{sec_intro}
Helping users in real-world manual tasks is a complex and challenging paradigm~\cite{Gottardi2022, wizard_of_tasks, taskbot_2023}, where it is necessary to leverage multiple information sources, provide several multimodal stimuli, and be able to correctly ground the conversation in a helpful and robust manner.
In this work, we build upon the success of TWIZ~\cite{twiz_2021} and introduce new features along with expanding existing ones. This results in an assistant capable of guiding a user through a task while keeping it engaging and stimulating.

With the aim of advancing Multimodal Conversational AI, we explore three main research questions encompassing several contributions:

\begin{itemize}
    \item \textbf{RQ1: Humanly-Shaped Conversations} - 
    Conversations should be fun and rewarding, yet reach a successful completion. To achieve this, we take an all-encompassing approach and propose novel ways of finding/creating a task, getting task highlights in the overview, and handling the task execution dialogue with TWIZ-LLM, a Large Language Model (LLM) trained specifically for supporting a robust interaction in the TaskBot domain.
    \item \textbf{RQ2: Multimodal Conversational Stimulus} - Interactions between a user and an assistant should make use of various stimuli to keep the conversation engaging. To this end, we leverage both text and visual content. Particularly, we expand the curiosities paradigm by generating more fun and contextual curiosities. On the visual side, we explore several image-generation methods and propose ways to make them more consistent with the target task. Finally, we present a novel Video Navigation feature, to allow for interactive video navigation by voice commands. 
    
    \item \textbf{RQ3: Zero-shot Conversational Flows} - Given that users are unpredictable and may not follow the expected interaction, we need a robust dialogue framework. The goal is to steer the user through a pleasant and natural conversation while supporting these conversation detours.
    For this, we leveraged various LLMs and prompting strategies to extend TWIZ-LLM to converse about unseen topics with answers that are both in-scope and meaningful to the conversation.
\end{itemize}

\section{TWIZ Bot Modular Architecture}
\label{sec_architecture}
\begin{figure}
    \centering
    \includegraphics[width=1.0\textwidth]{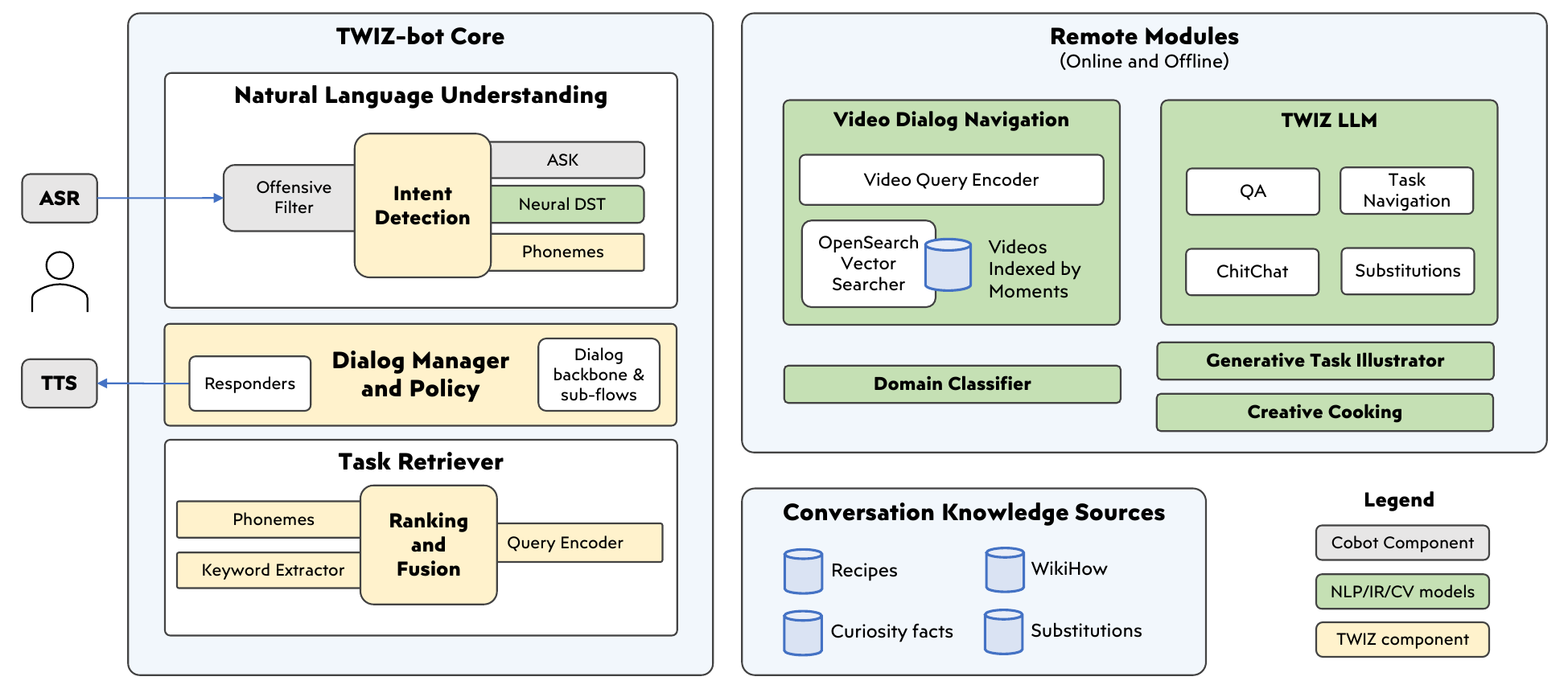}
    \caption{Architecture diagram of the TWIZ bot.}
    \label{fig_architecure}
\end{figure}

An overview of our architecture is presented in Figure~\ref{fig_architecure}. It is built upon Amazon's CoBot~\cite{cobot_2018}, a framework made available by Amazon for developing conversational agents. The agent operates within AWS Lambda, while all machine learning algorithms run on external modules, resulting in a highly efficient architecture with optimal performance. Leveraging the framework's layered architecture, we adopt a shared database pattern using Amazon's DynamoDB.

\subsection{Intent Detection}
\label{sub_intent_detection}
Understanding the user's intent is crucial to keeping a smooth flow in the dialogue interactions.
Given its success, we use an approach similar to the previous year~\cite{twiz_2021} by combining three methods: 1) phonemes-based matching; 2) rule-based corrections; and 3) a BERT-based model~\cite{tavares_task_2023} for intent detection.
These models work in an in-domain setting. However, they struggle to handle unseen or long-tail intents, requiring re-training to introduce new intents.
To tackle this, we introduce a zero-shot intent detection method (Section~\ref{sub_zero_shot_dst}).

\subsection{Dialogue Manager}
\label{sub_dialog_manager}
The dialogue manager keeps track of the current state and flow of the conversation.
Our dialogue manager,
adopts an event-driven state machine pattern. The progression through its different states is triggered by events associated with the detected intent of each user utterance, allowing for the context of a state to be used in order to provide relevant responses. With it, it is possible to keep track of the conversation's progress and leverage state transitions to maintain a correct conversation flow.

\subsubsection{Dialogue Backbone-Flow}
The dialogue manager provides graceful guidance to the user through the task at hand, by means of a \textit{backbone-flow}, which is a set of states that are necessary to go through in order to progress through a task. In TWIZ, the backbone states are as follows:

\textbf{1. Greeting} - The starting state, in which TWIZ's functionalities are presented and task suggestions are made (e.g., Summer suggestions).

\textbf{2. Grounding} - State in which the selected suggestions or search results (Section~\ref{sub_frictionless_search}) are shown to the user, so they can make a choice.

\textbf{3. Task Overview} - State after the selection of a particular task, where its overview -- e.g. rating, duration, and others -- is presented. 
This enables the user to either initiate the task or return to grounding in order to alter their selection.
We also add a generated description to provide a brief task summary and further entice the user to start the task  (Section~\ref{sub_task_promoter}).

\textbf{4. Task Execution} - The user can browse through the different steps of the task. Multimodal devices provide visual depictions of the steps, presented as either original or generated images (Section~\ref{sub_visual_dialog_exploration}), or videos that the user can interact with through touch and voice (Section~\ref{sub_moment_retrieval}).

\textbf{5. Task Completed} - This is the last state of the dialogue, after the execution of the task. The agent offers more task suggestions, and the user has the option to either select one of these tasks, conduct a new search, or conclude the interaction.

\subsubsection{Dialogue Sub-flows}
Subsequently, the dialogue \textit{backbone-flow} is enriched with \textit{sub-flows}, comprised of a set of states associated with an additional feature, such as answering questions and engaging in chit-chat (Section~\ref{sub_twiz_llm}), sharing curiosities (Section~\ref{sub_generated_curiosities}), and navigating through a video (Section~\ref{sub_visual_dialog_exploration}).

The defined sub-flows are organized in a self-contained fashion, allowing for easy insertion, modification, or removal. 
The possibility to seamlessly integrate or remove sub-flows or states without causing disruptions to the rest of the state machine allows different developers to focus on various features without having to worry about conflicts with states, associated events, and response generators.
The usage of different sub-flows also provides support for dialogue flexibility, allowing the user to stray from the main conversation in a contained way. By keeping track of the conversation’s current state, the dialogue manager employs a stack-like checkpoint mechanism to offer seamless fallback options, guiding the user back to the \textit{backbone-flow}.

While the primary objective of the dialogue manager is to assist the user in carrying out a task, it also accommodates the user's ability to switch tasks midway and provides the option to pause and resume the task at a later time, preserving all the progress made in that particular task.

\section{RQ1: Humanly-shaped Conversations}
\label{sec_humanly_shaped_conversations}
Making sure that users select and successfully complete a task is one of the main objectives of a TaskBot.
Therefore, the research goal is to deliver a conversation flow in an open, yet resilient manner. Our quest is to strike the delicate balance between the \textit{task-knowledge} and a \textit{humanly-shaped dialog} to ground the conversation on a complex manual task and guide the user through it.

\subsection{Frictionless Task Suggestions and Search}
\label{sub_frictionless_search}
TWIZ needs to swiftly ground the conversation in a frictionless way, by effectively and collaboratively finding the right task while also allowing for an exploration paradigm.
Consequently, on the home page, we present a range of demonstrative tasks for users to select. These examples not only provide a clear overview of TWIZ's capabilities, but also offer users insights into the type of supported tasks.
Moreover, from our analysis, a large portion of the time, users follow the suggestions provided, highlighting the value of these examples in guiding the users.
Expanding upon these insights, we incorporated time-sensitive suggestions, to further improve the relevance of the tasks shown, as well as seasonal tasks relevant to the current time of the year.
All the tasks suggested on the home page were manually selected to ensure a high-quality user experience.

The task search pipeline follows a multi-query, multi-ranking, and rank fusion approach.
First, user queries are processed and the most relevant terms are extracted~\cite{twiz_2021}. Secondly, we apply a ranking algorithm. 
Our approach involves carrying out both lexical (text) and semantic (embedding) searches on an OpenSearch index, with semantic searches conducted using embeddings generated by the MPNet model~\cite{mpnet}.
The results are then combined and re-ranked using the cosine similarity against the user query, and an additional step of heuristic-based task quality parameters, such as the presence of a video or the number of ratings. This two-step approach enhances our ability to deliver the most relevant and quality-assured tasks.

\subsection{Creative Cooking: \textit{What's in your fridge?}}
\label{sub_creative-cooking}
With regards to cooking, a helpful TaskBot should be able to guide users through cooking deadlocks (e.g. missing an ingredient, no matching recipe).
To this end, we introduce a novel \textit{Creative Cooking} feature that seeks to push the limits of users' creativity, TWIZ's knowledge, and cooking. In particular, it seeks to boost the users' creativity by letting them create their own unique and personalized tasks, adjusted to their preferences and available ingredients.

For example, if the user has in the fridge a particular set of ingredients (e.g. zucchini and eggplant) and a particular cooking style in mind (e.g. vegetarian) the user should be able to prompt TWIZ with a recipe that has these particular constraints, resulting in a recipe such as "grilled veggies".
To do this, we first extract ingredients and cooking styles from the user utterances, using a rule-based approach against a set of curated tags. This way, we can guarantee that only valid tags are used.
After this, we provide the user with a set of recipes from the API that satisfies the user's requests, and we add a recipe generated using an LLM considering the user's specifications.
In particular, we use a Vicuna~\cite{vicuna2023} model, which we prompt to generate a title, a list of ingredients, and a list of steps, given the aforementioned set of tags.
An example of the creative cooking feature can be seen in this \href{https://www.youtube.com/playlist?list=PLC5saXed4eNtMDJPITQ0M4i0SGD83k_gy}{video}\footnote{\url{https://www.youtube.com/playlist?list=PLC5saXed4eNtMDJPITQ0M4i0SGD83k_gy}}. In Figure~\ref{fig_creative_cooking}, we show an example of the creative cooking feature with a generated recipe, illustrated with our image generation methods (Section~\ref{sub_visual_dialog_exploration}).

The creative cooking feature seamlessly integrates with all other TWIZ features (e.g. curiosities and image generation pipeline).
As future work, we aim to create a user study for comparing manually created to generated recipes.

\subsection{Task Overview: Task Promoter}
\label{sub_task_promoter}
When analyzing our interactions, we noticed that there is a direct correlation between users who start a task and higher ratings (Section~\ref{sub_user_behavior}).
With this in mind, we looked for new ways to entice users to start a task during the \textit{Task Overview} phase.
Consequently, in the previous year, this led to the development of a 3D visual illustrator of the recipe~\cite{twiz_2021} and manual templates to highlight certain features of a recipe~\cite{twiz_2021}, which resulted in low diversity responses.
This year, we developed a Task Promoter, whose purpose is to generate, for any given task, a short and appealing description that highlights the best it has to offer.

\subsubsection{Evaluation}
\label{sub_sub_promoter_human_eval}
We conducted a human evaluation, considering the recipes domain, and manually assessed for 100 recipes the preferred description: a prompted Vicuna-7B~\cite{vicuna2023} or a fine-tuned GPT-2~\cite{radford_gpt2} based model which we call RePro (for model training details refer to Appendix~\ref{app_promoter_training_details}). 
Additionally, we ask annotators to identify non-sensical descriptions and/or comprising ingredient hallucinations. 
To help assess the hallucinations, we provide annotators with both the recipe name and the ingredients, as in Table~\ref{tab_promoter_examples}. 
We collect two annotations for each pair of descriptions. 

\begin{table}[htbp]
\centering
\begin{tabular}{lcccc} \toprule
  \textbf{Model}     & \textbf{Win \%} & \textbf{Hallucination\%} & \textbf{Broken \%} & \textbf{\# Params} \\ \midrule
RePro (GPT-2)  & 20 & 57 & \textbf{10} & 770 M \\
Vicuna-7B & \textbf{61} & \textbf{29} & 11 & 7000 M \\
\bottomrule
\end{tabular}
\vspace{2mm}
\caption{Results of manual evaluation of the recipe promoter. Ties are not included in \textit{Win \%.}}
\label{tab_repro_test}
\end{table}

The results in Table~\ref{tab_repro_test} show that Vicuna in a zero-shot scenario resulted in preferred generations over a purpose-built smaller model. Although both models present very few non-sensical generations, a key differentiator is the hallucination of ingredients in the recipe, where over half of the RePro descriptions had at least one hallucinated ingredient. 
However, we noticed that almost all hallucinated and broken descriptions are easily identified, allowing them to be automatically removed.

Given the success of the Vicuna-7B approach, we expanded this feature to the WikiHow domain in a similar way using the title of the task.
To improve model performance, as a next step, we plan to train a \textit{RePro} model based on outputs from an LLM such as GPT-4.

\subsection{Task Execution: TWIZ LLM}
\label{sub_twiz_llm}
One of our team's major focuses this year is to make TWIZ more natural and robust to user dialogues. Due to the emergence of effective LLM-based chatbots, users have higher expectations when interacting with TWIZ. To meet these expectations, we developed a task-oriented LLM-based approach, that seeks to   
support the \textit{Task Execution} phase\footnote{When the user has already selected and started a task.}.
We focused only on recipes, but plan to further expand it to DIY tasks.

\subsubsection{Conversational Data Augmentation}
\label{sub_sub_llm_data_creation}
During the task execution phase, we want to allow users to fully explore the recipe rather than just advance to the next step, to be able to ask questions about the cooking process, replace an ingredient, and get advice on the tricky parts. To generate data that accounts for all of these sub-flows in a robust manner, we devised a dialogue-generation pipeline that leverages data gathered during our participation in the first edition of the Alexa TaskBot Challenge~\cite{Gottardi2022}.
We complement this data with several data augmentation techniques, described below, to increase the diversity of the generated dialogues.

\paragraph{Conversation Flow}
The basis of the dataset is the policy that dictates how dialogues are generated. For this, we extracted user patterns from user interactions and created a directed graph containing all the identified intents and the probabilities of the user transitioning between them, allowing us to accurately model real-user behavior when generating a new dialogue. The intents considered are classified using a combination of rule-based heuristics and a customized Transformer-based intent classifier~\cite{twiz_2021}. The complete list of intents considered and their description can be seen in Table~\ref{tab_intents_list}.
To improve dialogue diversity, we manually increased the transition probabilities of less common intents such as Ingredient Replacement and Questions. While, by doing this, we are making the generated dialogues policy diverge from the real-user policy, we believe the added frequency of such intents can greatly improve the ability of models trained on this data to attend to this type of intents. 

As this dataset focuses on task execution, we do not consider the task selection and search phases and simulate the generated dialogues starting when the user asks the assistant to start an already selected task. 
An example of a generated dialogue can be seen in Appendix Table~\ref{tab_example_conversation}.

\begin{table}[htbp]
\centering
\small
\resizebox{\textwidth}{!}{%
\begin{tabular}{clllc}
\toprule
 & \textbf{Intent}        & \textbf{Description}                  & \textbf{Example} & \textbf{\# ex.}\\ \midrule
\multirow{7}{*}{\rotatebox{90}{\textbf{\ \ \ Navigational}}} 
 & Next Step              & Go to the next step of the task.      & Next Step.       & 169                      \\
 & Repeat                 & Repeats the last system utterance.    & Repeat that.     & 76                     \\
 & Stop                   & Ends the interaction.                 & Stop             & 73                    \\ 
 & Yes                    & Confirmation (acts as Next Step).     & Sure.            & 52                      \\
 & Previous Step          & Go to the previous step of the task.  & Previous Step.   & 21                     \\
 & Resume                 & Repeats the current step.             & Resume.          & 18                     \\
\midrule
\multirow{4}{*}{\rotatebox{90}{\textbf{Question}}} 
 & Get Curiosities               & Asks for a fun fact.           & Can you tell me a fun fact? & 13 \\ 
 & Ingredients Replacement       & Asks for a replacement.        & I do not have sugar.        & 9         \\
 & Definition Question           & Definition type question       & What is a spatula?          & 4         \\
 & Question                      & Task specific question.        & How much salt do I need?    & --         \\
 \midrule
\multirow{5}{*}{\rotatebox{90}{\textbf{Other}}}
 & Fallback         & Intent is not recognized.             & Find a restaurant near me.         & 2618   \\
 & Sensitive        & Mentions a sensitive/dangerous topic. & How do you make a nuke?            & 235  \\
 & Chit-Chat        & Chit-chat utterances.                 & How are you today?                & 131   \\
 & Search           & Ask for another task.                 & How to change a tire?             & 108   \\
 & More Detail      & Ask for more details about a step.   & More details, please.             & 36     \\ 
 \bottomrule
\end{tabular}%
}
\vspace{2mm}
\caption{Intents list, description, example (manually-crafted) utterances, and the count of unique utterances available. }
\label{tab_intents_list}
\end{table}

\paragraph{Task Selection} As, in each dialogue, the user focuses on completing a single recipe. We used Amazon's provided recipes dataset and extracted 1000 recipes that had between 5 and 10 steps, and a total word count of no more than 350. This helps ensure that the recipes are long enough for a meaningful dialogue, with opportunities to ask questions, while also avoiding overly long recipes that could lead to repetitive or noisy dialogues.

\paragraph{User Queries} For the data to closely mimic real user behavior, we used authentic user requests. To achieve this, we collected all user utterances classified for each considered intent and their absolute frequency. To clean up classification errors, we manually reviewed each of the most common user requests for each intent and removed any utterances that did not match the intent. When generating a new dialogue turn, the user utterance is selected from the list of utterances for the current intent using a random weighted selection, where the weight of each utterance is its absolute frequency, ensuring that more common utterances are more likely to be selected.
However, some intents are extremely contextual, meaning that their associated user requests can have vastly different answers based on the relevant recipe (e.g., questions regarding a specific recipe step or ingredient replacement). 
For these cases, we utilize zero-shot LLM prompting and template-based approaches to produce natural-sounding and context-relevant user utterances. 
More information about this can be found in Appendix~\ref{app_twiz_llm_dataset_creation}.

\paragraph{System Responses}
The system responses need to be diverse but also accurate and contextual, w.r.t. the recipe and the user request. To achieve this, based on the user intent, we use templates, knowledge bases, and LLM-prompting:
\begin{itemize}
    \item \textbf{Navigational} - For these cases the system response consists only of step text.
    \item \textbf{Definition Question} - We query a dictionary\footnote{\url{https://github.com/wordset/wordset-dictionary}} for the appropriate meaning of a given concept.
    \item \textbf{General Question} - Given a recipe, generate QA-pairs by prompting OpenAI's \textit{text-davinci-003}. This way both the questions and system answers are contextual. 
    \item \textbf{Ingredient Replacement} - Template responses, filled in using a knowledge base\footnote{\url{https://foodsubs.com/}} of valid ingredient replacements (e.g., ``\{Ingredient A\} can be replaced with \{Ingredient B\} in this recipe''). 
    \item \textbf{Fun facts} - We used OpenAI's \textit{text-davinci-003} to generate curiosities for each recipe step by providing the step text and a relevant Wikipedia paragraph\footnote{using \url{https://neuml.github.io/txtai/}} and prompting the model to generate a fun fact relevant to that step.
\end{itemize}
For all other cases, a template-based approach was used. The templates were written by hand with up to 5 examples per case. 
In these cases, to ensure inter-dialogue response diversity, the selected response was randomly sampled from all templates not used in the past 5 dialogue turns. Additionally, every template is written in several different tones of voice, to allow for more diversity amongst the generated dialogues.

\paragraph{Dialogue Generation Pipeline}
The complete generation pipeline follows the following three steps when generating a new dialogue turn:
\begin{enumerate}
    \item \textbf{Determine User Intent} - Based on the previous turn and on the extracted policy, sample the next user intent.
    \item \textbf{Retrieve User Utterance} - Based on the selected intent and current recipe step, retrieve a candidate user utterance.
    \item \textbf{Produce System Response} - Based on the selected intent, user utterance, and recipe, select the appropriate system response.
\end{enumerate}
These steps are repeated until the user reaches the last step of the recipe or a \textit{Stop} intent is selected.

\subsubsection{Base LLM Models}
\label{sub_sub_llm_model}
We are currently using two models: \textbf{Vicuna-7B}~\cite{vicuna2023}, a LLaMA-based~\cite{llama} model fine-tuned on conversational data, and \textbf{OPT-1.3B}~\cite{opt}, to understand how different model sizes and architectures behave on this task.

For the model input, there are four key pieces of information that the model needs to be able to attend to (training prompts are shown in Appendix~\ref{app_example_prompts}):
\begin{itemize}
    \item \textbf{Tone of Voice:} The tone that should be used by the system during the dialogue. It can be neutral, somewhat polite, polite, or very polite. This allows training models with controllable tone of voice in their responses.
    \item \textbf{Recipe Text:} The recipe title, followed by its steps. 
    \item \textbf{Current Step:} The recipe step that the user is currently on. This helps in navigational requests and and in keeping the answers grounded.
    \item \textbf{Dialogue context:} The previous $t$ turns + the \textit{current user request}. For our evaluation, we use a $t=1$.
\end{itemize}

\subsubsection{Experimental Setup}
\label{sub_sub_llm_imp_details}

\paragraph{Model Details} We trained Vicuna-7B and OPT-1.3B on the same data. Vicuna was trained for 1 epoch, whereas OPT was trained for 10. In addition to SFT training, we also trained OPT under an RLHF paradigm~\cite{sumarize_rlhf,instructgpt}, to understand if the models would benefit from this approach.

\paragraph{Dataset} We generated 10k dialogues with an 80/10/10 split. 
To create the negative system responses for RLHF, we employ different methods based on the user's intent: 
\begin{itemize}
    \item \textbf{Navigational} - we introduce a sentence from the previous/next step to simulate model copying mistakes.
    \item \textbf{Definition Questions} - we select the definition ranked lowest using a bag-of-words approach.
    \item \textbf{Ingredient Replacement} - we randomly choose an ingredient from a list to replace the current one.
    \item \textbf{Get a Curiosity} - we retrieve a curiosity with lower similarity using a bag-of-words approach.
    \item \textbf{Sensitive} - provides an answer to a sensitive request given by an uncensored LLM.
\end{itemize}
If no change is applied, and for all other intents, we introduce response perturbation and grammatical errors.

\subsubsection{Automatic Evaluation}
\label{sub_sub_llm_results}

\begin{table}[tbhp]
\centering
\resizebox{\textwidth}{!}{%
\begin{tabular}{l|c|cc|cc|cc} \toprule
                        & \multirow{2}{*}{Test Set \%} & \multicolumn{2}{c|}{OPT-SFT} & \multicolumn{2}{c|}{OPT-RL} & \multicolumn{2}{c}{Vicuna} \\
                        &                                         & METEOR            & BScore-F1    & METEOR           & BScore-F1    & METEOR           & BScore-F1   \\ \midrule
Navigational            & 65.36\%                                 & 64.39             & 81.73            & 59.79            & 79.90            & \textbf{98.36}   & \textbf{99.29}  \\ \midrule
Get Curiosities         & 4.83\%                                  & 14.94             & 57.44            & 14.52            & 57.11            & \textbf{18.29}   & \textbf{61.96}  \\
Ingredients Replacement & 1.11\%                                  & 77.37             & 87.59            & 32.92            & 68.84            & \textbf{79.91}   & \textbf{89.19}  \\
Definition Question     & 4.07\%                                  & \textbf{83.76}    & \textbf{91.52}   & 83.09            & 91.23            & 81.25            & 91.23           \\
Question                & 9.41\%                                  & 54.01             & 78.52            & 53.26            & 78.20            & \textbf{61.80}   & \textbf{82.92}  \\ \midrule
Sensitive               & 4.07\%                                  & 81.75             & 90.83            & \textbf{91.00}   & \textbf{95.30}   & 68.30            & 85.52           \\
Other*                  & 11.15\%                                 & 68.97             & 83.57            & 42.66            & 73.06            & \textbf{84.78}   & \textbf{92.65} 
\\ \bottomrule
\end{tabular}
}
\vspace{2mm}\caption{Per intent results for all three trained models. *Sensitive intent was separated from the Other intent group due to outlier results.}
\vspace{-5mm}
\label{tab_per_intent}
\end{table}

To measure model performance, we evaluate the model on all turns of the test set dialogues. 
For automatic metrics, we considered METEOR~\cite{meteor} and BERT-Score-F1~\cite{bertscore} to measure textual overlap and semantic similarity, respectively.
In Table~\ref{tab_per_intent}, we provide a detailed analysis of the models' performance, by separating turns with the ``more static'' intents, which require copying from the task information (e.g., \textit{Navigational}) or returning a default response (e.g., \textit{Other}), from the turns that require access to some external knowledge. 

These early results show that the Vicuna-based model is capable of handling \textit{Navigational} intents. It also outperforms both OPT-based models for almost all intents. 
For Vicuna, the outliers are the \textit{Curiosity} requests. These answers are difficult to evaluate using automatic metrics since there can be various correct answers given a user's request.
RL on the OPT model seems to have little impact in most cases. However, it does improve \textit{Sensitive} requests, while having a negative impact on both \textit{Ingredient Replacement} and on the \textit{Other} intent group, indicating that the model might confuse these requests as dangerous tasks. We also noted that when handling sensitive requests, both OPT and Vicuna recognized the nature of the request and did not respond inappropriately.

\subsubsection{Human Evaluation}
\label{sub_user_evaluation}
To more accurately measure the models' performance, we conducted a human evaluation for Ingredient Replacement, Questions, and Curiosity user requests. We focused on these request types due to their reliance on external knowledge and the possibility of multiple correct answers.
For the 3 cases, we selected 50, 100, and 30 examples from the test set, respectively. The annotators were asked to rate each one on a scale of 0 (wrong/irrelevant) to 2 (accurate/very relevant). For both ingredient replacement and questions, the criteria used was accuracy, whereas for curiosity requests, the relevancy of the generated curiosity w.r.t. the current recipe step was annotated. 
Annotators were instructed to rate 0 in cases where the generated utterance is incoherent or a clear hallucination.

Table~\ref{tab_human_evaluation} shows the results of the human evaluation.
Vicuna outperforms both OPT models. Using RL on OPT had a negative impact on model performance on two tasks, and improved for \textit{Question} requests. For \textit{Curiosity} requests, OPT produces fairly relevant curiosities, on par with Vicuna, but performance worsens with the RL approach.

\begin{table}[htbp]
\centering
\begin{tabular}{@{}lccc@{}}
\toprule
\textbf{Model}                  & \textbf{OPT-SFT} & \textbf{OPT-RL} & \textbf{Vicuna} \\ \midrule
Question               &           1.30           &         1.48             &         \textbf{1.77}       \\
Ingredient Replacement &           0.84           &         0.76             &         \textbf{1.48}        \\
Get Curiosities        &           1.23           &         0.90             &         \textbf{1.30}        \\ \bottomrule
\end{tabular}%
\vspace{2mm}
\caption{\label{tab_human_evaluation}Human evaluation results of TWIZ LLM models on the Question, Ingredient Replacement, and Fun Fact intents, on a 0 to 2 scale.}
\end{table}

In a later analysis of the reward model used with OPT-RL, we noted that the reward model did not learn to model the preferences appropriately.
We believe this is due to the low quality of the preference data. 
However, these results, along with the results observed in automatic evaluation, show that RLHF has the potential to provide meaningful performance improvements but lacks consistency. 
Thus, future work should focus on improving the quality of the generated preference data, for example, using more diverse LLM prompting methods.

To conclude, TWIZ-LLM presents a step forward in having a model capable of guiding a user through the execution of a task while providing natural and helpful responses.

\section{RQ2: Multimodal Conversation-Stimulus}
\label{sec_multimodal_stimulus}
Conversation stimulus is what advances a conversation, i.e., the user's desire to obtain the final outcome of the task and the road that leads to it.
TWIZ provides a number of linguistic, visual, and cognitive stimuli to keep conversations natural and engaging.

\subsection{UI and APL Templates}
\label{sub_apl_templates}
The user interface significantly influences the perceptions and experiences of the users. As such, the quality and functionality of the interface are crucial. To address this, we developed a new user interface that is more capable of showcasing TWIZ's features.

\begin{figure}[t]
     \centering
     \begin{subfigure}[b]{0.32\textwidth}
         \centering
         \includegraphics[width=\textwidth]{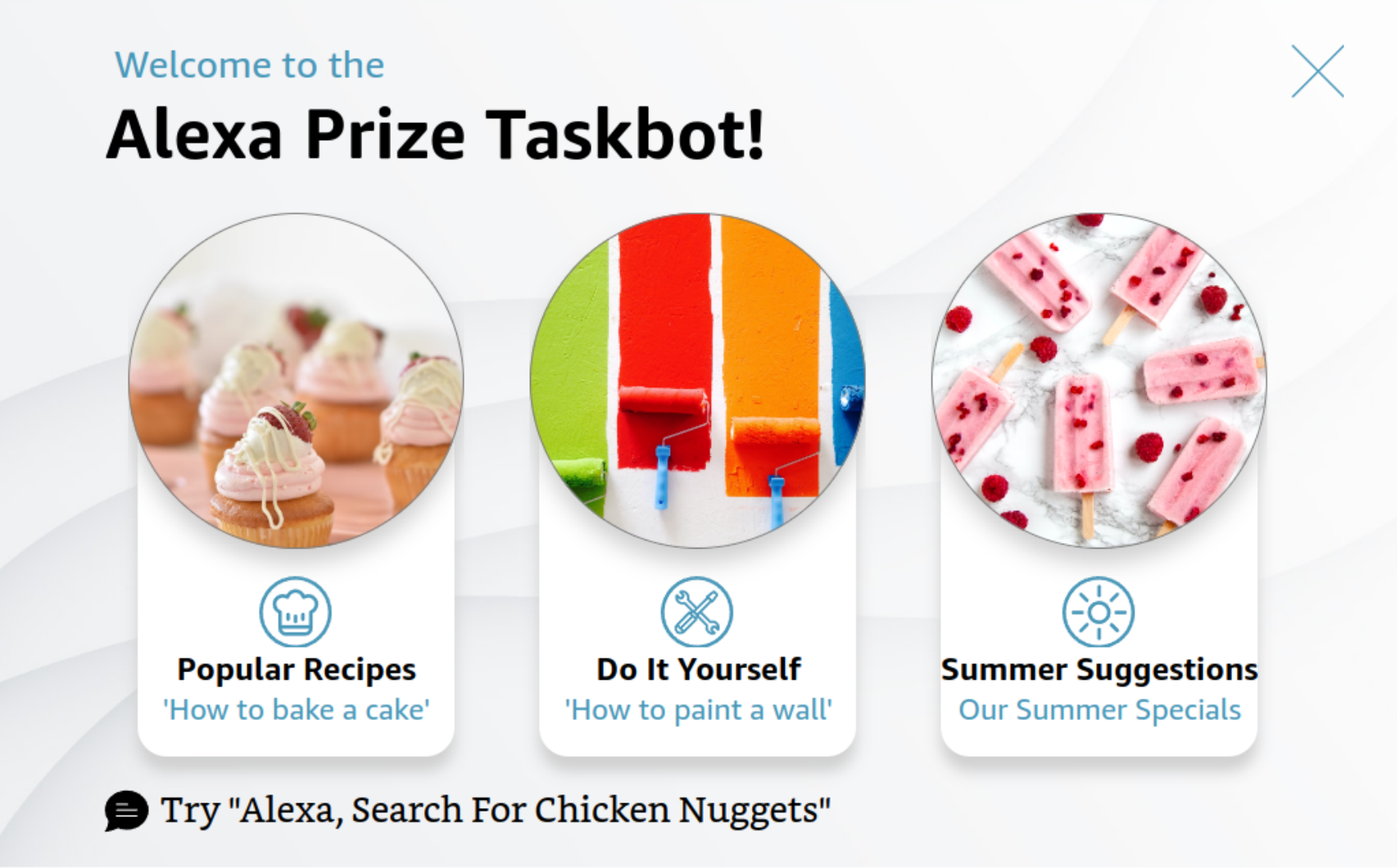}
     \end{subfigure}
     \hfill
     \begin{subfigure}[b]{0.32\textwidth}
         \centering
         \includegraphics[width=\textwidth]{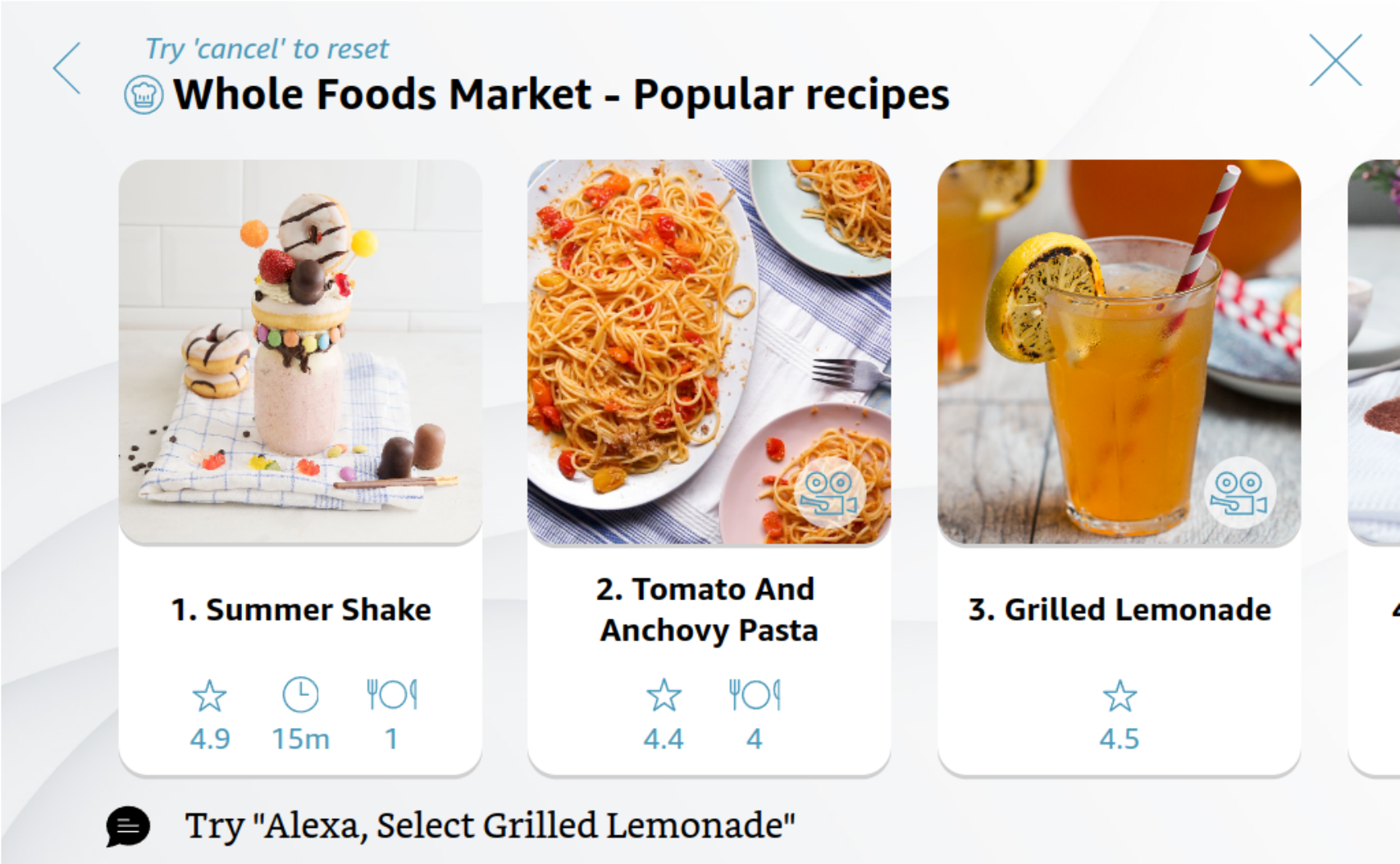}
     \end{subfigure}
     \hfill
     \begin{subfigure}[b]{0.32\textwidth}
         \centering
         \includegraphics[width=\textwidth]{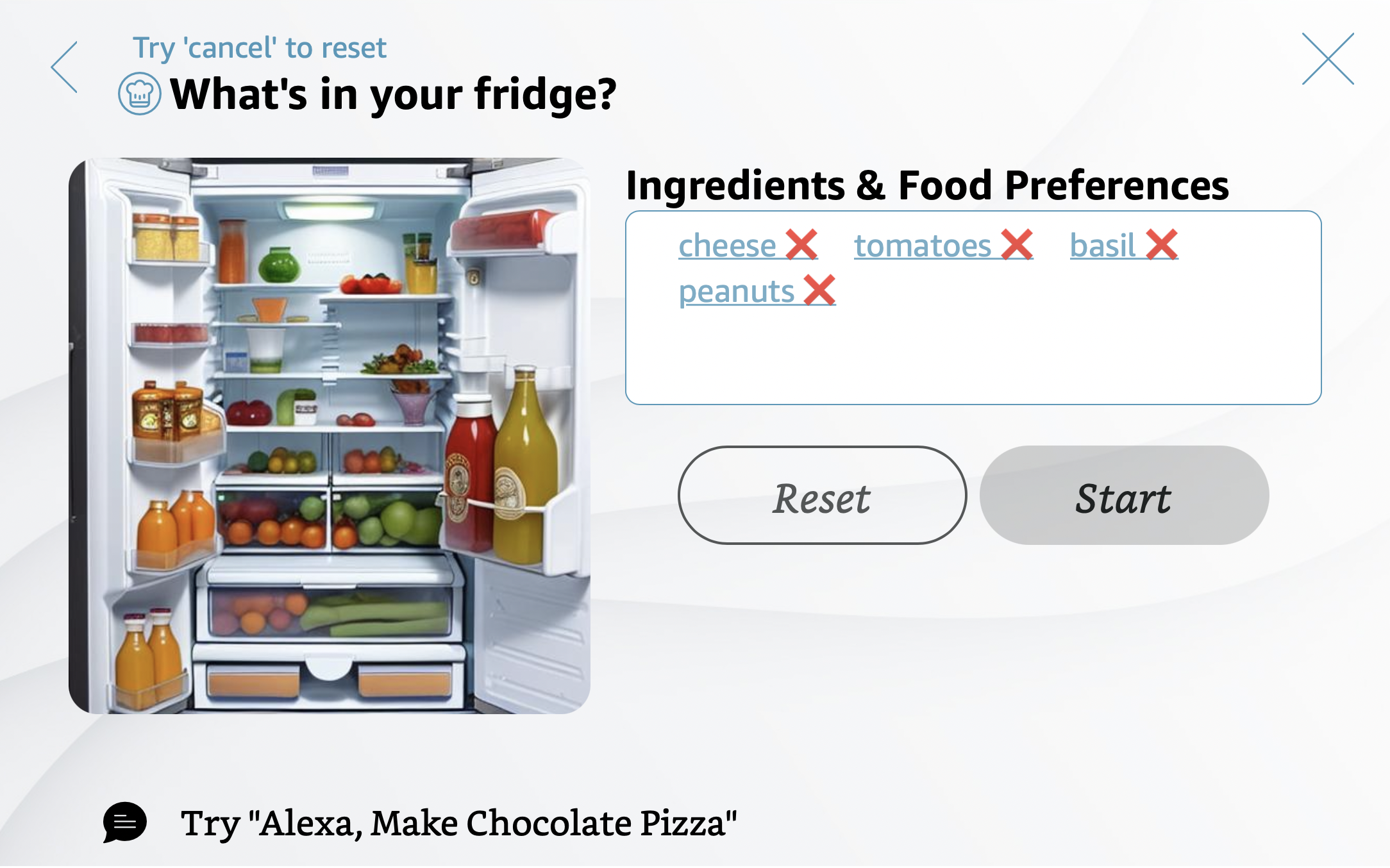}
     \end{subfigure}
     \vspace{2mm}

     \begin{subfigure}[b]{0.32\textwidth}
         \centering
         \includegraphics[width=\textwidth]{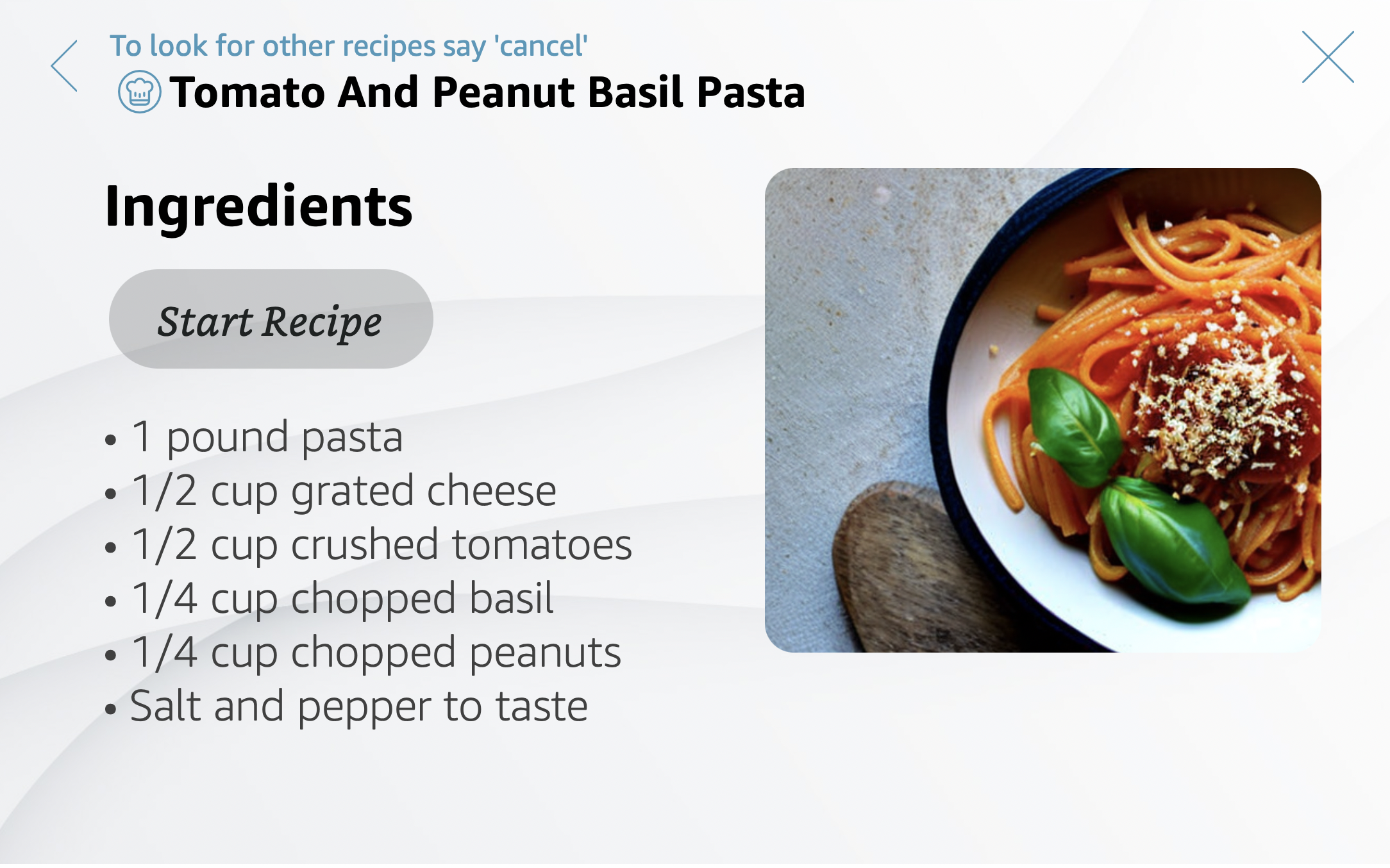}
     \end{subfigure}
     \hfill
     \begin{subfigure}[b]{0.32\textwidth}
         \centering
         \includegraphics[width=\textwidth]{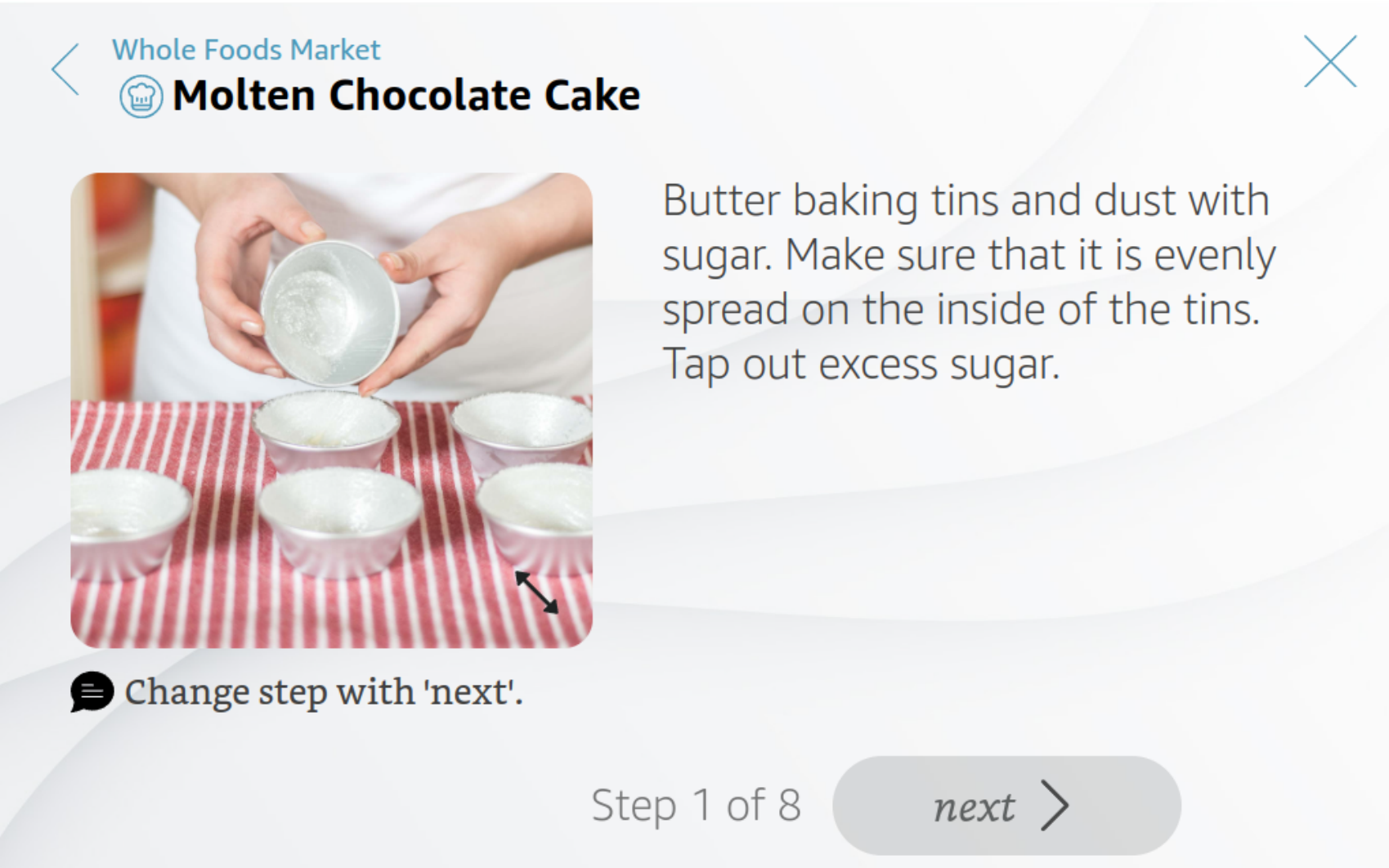}
     \end{subfigure}
     \hfill
     \begin{subfigure}[b]{0.32\textwidth}
         \centering
         \includegraphics[width=\textwidth]{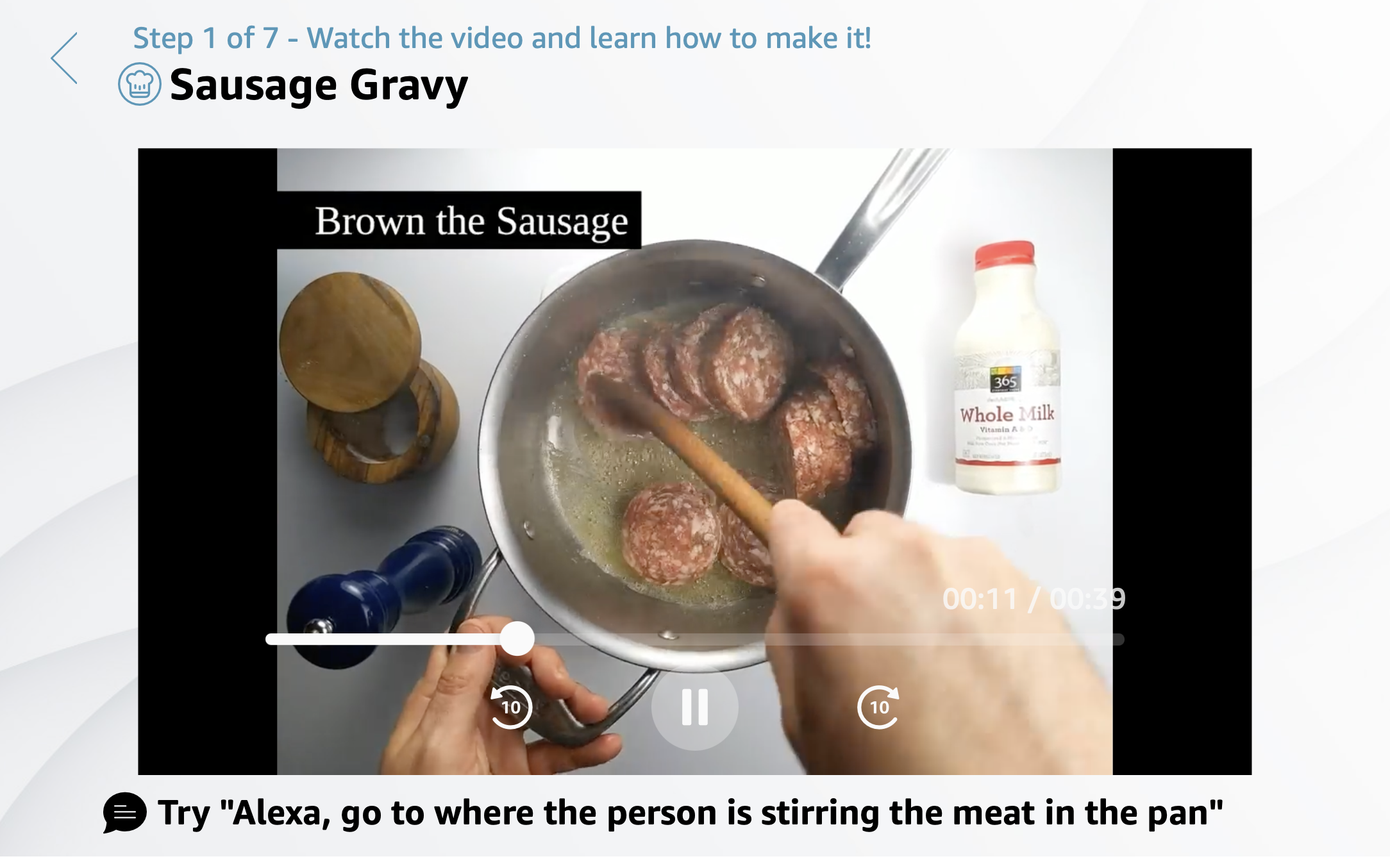}
     \end{subfigure}
     \vspace{2mm}
     \caption{TWIZ's APL screens: \textit{welcome screen}, \textit{search results}, \textit{``What's in my fridge?''}, \textit{ingredients list}, \textit{task step}, and \textit{video dialogue}.
     }

     \label{fig:apl}
\end{figure}

Our design philosophy leans towards minimalism, emphasizing a clean theme to promote clarity and ease of use. This approach encourages a seamless user experience, contributing to more efficient interactions.
Additionally, on each screen, we provide a rotating set of contextual tips that introduce users to TWIZ's wide range of features.
The complete set of screens can be found in Figure~\ref{fig:apl}.

\subsection{Task Visual Enrichment and Dialogue}  
\label{sub_visual_dialog_exploration}
Task illustrations are fundamental for rich and engaging visual-stimuli interactions. Nonetheless, a significant number of tasks either lack accompanying illustrations or have low-quality ones. Moreover, creative cooking (Section~\ref{sub_creative-cooking}), offers the possibility to compose a recipe from scratch, requiring a strategy to illustrate it.
In this section, we describe how image generation models~\cite{stable_diffusion}, can provide a solution to generate task and step-specific illustrations.

\subsubsection{Generative Task Illustration}
\label{sub_sub_task_illustration}
Illustrating a task with text-to-image algorithms can be challenging, and requires extra care to ensure the generated images are grounded on the task and have good quality. Our implementation can be divided into three main steps:
\begin{enumerate}
    \item \textbf{Image generation} is a non-trivial task, as it involves generating images from a task that might not be visual in nature.
    We propose a new method~\cite{nl2vi}, which uses an LLM to convert the task into a visual prompt with high consistency 
    that can then be used in text-to-image models.
    We generate multiple images for each task using the Stable Diffusion XL model~\cite{stable_diffusion_xl}. 
    \item The \textbf{image score} is given by an alignment score.
    This alignment score is calculated by the NL2VI metric~\cite{nl2vi}, 
    which measures the alignment of the image to the prompt. The alignment is calculated using VQA algorithms~\cite{blip-1, mplug, ofa, git}. The questions for VQA are generated using an LLM with in-context learning and are filtered using question answering~\cite{unifiedqa} and NLI~\cite{nli-model} models. This score ensures that the image is grounded on the task. The results of Table~\ref{tab:NL2VI-results} show that our method achieves better results than the TIFA~\cite{tifa} baseline.
        \item Lastly, we \textbf{rank the generated images}, as well as the original image, according to their score and choose the one with the highest score to illustrate the task~\cite{nl2vi}. 
\end{enumerate}

\begin{table}[tbp]
    \centering
    \begin{tabular}{@{}lccc@{}}
    \toprule
        & \textbf{Equality} & \textbf{NLI} & \textbf{BERT-Score} \\ \midrule
    \multicolumn{2}{l}{\textbf{TIFA}~\cite{tifa}} \\\midrule
    \quad Recipes & 64.7 & 68.9 & 72.0 \\
    \quad WikiHow & 56.1 & 61.6 & 64.2 \\
    \midrule
    \multicolumn{2}{l}{\textbf{NL2VI}~\cite{nl2vi}} \\\midrule
    \quad          Recipes & 73.7 & 76.2 & \textbf{79.8} \\
    \quad          WikiHow & 73.1 & \textbf{75.8} & 73.8 \\
    \bottomrule
    \end{tabular}
    \vspace{2mm}
    \captionof{table}{Comparison of the image alignment score between our method NL2VI~\cite{nl2vi} and the previous SOTA TIFA~\cite{tifa}. 
    }
    \label{tab:NL2VI-results}
\end{table}

\subsubsection{Generative Recipe Step Illustration}
\label{sub_sub_recipe_step_illustration}
After having a method to illustrate the main tasks' content, we now want to illustrate every step of a task using an image-generation approach, with visual continuity guarantees, which we refer to as \textit{Coherent Step Illustration}.
As established in Section~\ref{sub_sub_task_illustration}, generated images must be consistent with the prompts, ensuring the desired elements are present in the image. 
This is important in task step illustration, as the illustrations must be faithful to the step to steer the user in the right direction.
The main focus of this method is the sequential nature of step illustrations. Unlike \textit{Task Illustration}, which is independent, \textit{Step Illustrations} have dependencies regarding the previous steps.

Our method takes as input a sequence of steps $\{S_1, S_2, ..., S_n\}$ and generates each image, $I_n$, which illustrates step $S_n$ by considering all previous steps. The motivation is to increase the coherence of the sequence of images. The method comprises three phases:
\begin{enumerate}
    \item \textbf{Sequence of Captions.} When generating an illustration for a recipe step, we want the illustration of step $S_n$ to consider the previously generated images more directly.
    Hence, to address this limitation, we propose to consider the captions~\cite{instruct_blip} of the previously generated images, instead of the text from previous steps.
    With these captions, our updated input can be written as, $\{C_1, C_2, ..., S_n\} \rightarrow I_n$.
    
    \item \textbf{Sequence-to-Prompt LLM.} To transform the sequence $\{C_1, C_2, ..., S_n\}$ into an appropriate image generation prompt, we train an LLM with sequences of (captions$+$step) and the target prompt.
    
    \item \textbf{Sequence-to-Prompt Generation.}  After training, we can generate the prompts by iteratively prompting the LLM with the current step and the previous captions. These generated prompts contain enough information to increase the coherence of the generated images, with respect to the previous generations.
    \end{enumerate}

Given the nature of the task, we only apply this method to the recipes domain.
In future work, we will conduct a user study to measure the performance of this approach.

\subsection{Video Moment Retrieval}
\label{sub_moment_retrieval}

\begin{figure}
    \centering
    \includegraphics[width=\textwidth]{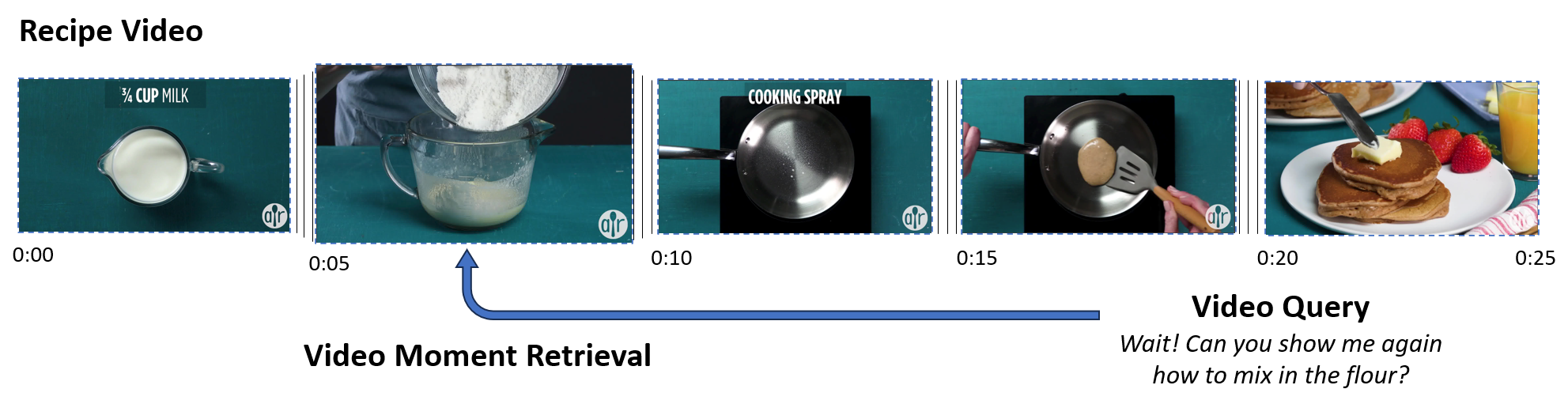}
    \caption{TWIZ users can ask a question about a video and navigate to the correct video moment that answers the question. An example interaction can be seen in this \href{https://www.youtube.com/playlist?list=PLC5saXed4eNsebM8C4W5S_BQ9ADEgwH57}{demo video link}.}
    \label{fig_vidmomret_example}
\end{figure}

In order to facilitate the interaction with the vast amount of information present in a video, we developed a method to 
enable users to control videos using voice commands
, when the video is in full-screen mode.
This allows users to interact with videos effortlessly and intuitively, enhancing their overall viewing experience, as in Figure~\ref{fig_vidmomret_example}.
An example interaction can be seen in this \href{https://www.youtube.com/playlist?list=PLC5saXed4eNsebM8C4W5S_BQ9ADEgwH57}{video}\footnote{\url{https://www.youtube.com/playlist?list=PLC5saXed4eNsebM8C4W5S_BQ9ADEgwH57}}.

\begin{wrapfigure}{r}{0.5\textwidth}%
  \vspace{-4mm}%
  \begin{center}%
    \includegraphics[width=0.48\textwidth, trim={0.5cm 0 1cm 1.1cm},clip]{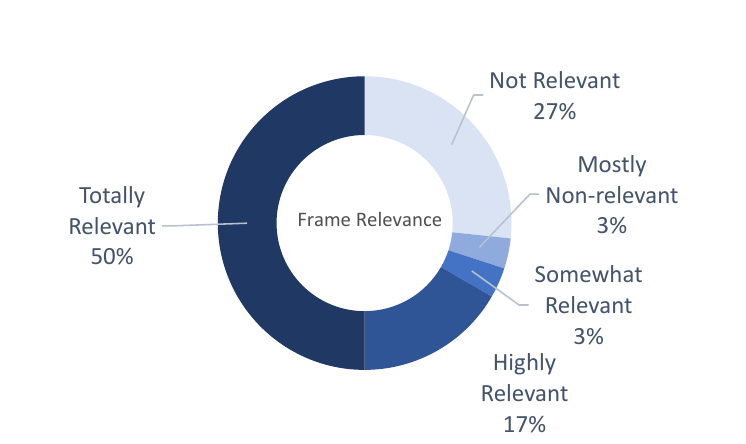}
    \vspace{5mm}
    \caption{Video moment retrieval results.\label{fig_nav_video_pie}}
  \end{center}%
  \vspace{-5mm}%
\end{wrapfigure}
As a first step, we extracted keyframes from all videos. 
For each keyframe, we generated a caption, using InstructBLIP~\cite{instruct_blip}. Then we used the image and caption of each frame to extract embeddings using CLIP~\cite{clip_model}.
All of this information is stored in an OpenSearch index, along with the task metadata.
The search is performed using the user query (e.g. ``\textit{When did the chef mix in the flour?}''), and uses both a full-text and embedding-based (text and image) search. 

The result thus has three sets of frames, one for each type of embedding (text and image) and one for plain text. 
We then perform re-ranking of the potential frames based on rank fusion. 
After selecting the winning frame, we perform a seek in the video based on the timestamp of the frame on the top of the rank.

To evaluate the proposed method, we ran a test of 30 queries over 10 videos.
For each user query, a video frame is returned. We judged the frame's relevance on a scale from 1 to 5, with 1 indicating a non-relevant frame and 5 indicating a totally relevant frame. As shown in Figure \ref{fig_nav_video_pie}, 27\% of the results were not relevant to the query, while 50\% were considered totally relevant.
Given these results, it is important to study ways to reduce the occurrences of lower-quality results by making better use of video information and improving the ranking algorithm.

\subsection{Fact-Grounded Curiosity Generation}
\label{sub_generated_curiosities}
Curiosity-stimuli, i.e. \textit{fun-facts}, turn conversations into an educational and memorable experience.
In the first edition of the TaskBot Challenge, we contextually enriched dialogues with a manually curated set of curiosities~\cite{twiz_2021}. 
Subsequently, we performed an A/B test with over 1000 conversations, which showed that curiosities increase user engagement and can also provide an average rating improvement~\cite{frederico_curiosities}.

Given these encouraging results, this year we expand this paradigm by using LLMs to generate more contextual and interesting curiosities, removing the bottleneck of manually creating curiosities.
In particular, we use GPT-4 as the backbone for our curiosity generation model.
Given that LLMs have a tendency to hallucinate~\cite{survey_hallucination}, we add a relevant Wikipedia passage as context to ground the generation, alongside the task title and current step.
We conduct a user study comparing the manually annotated curiosities with the generated ones, according to the following aspects: (1) \textbf{relevance} to the step, (2) \textbf{truthfulness} according to the provided information, and (3) \textbf{fun factor}. These aspects are measured on a scale of 0 (low) to 2 (high). Additionally, we also ask annotators to choose a winner between the two methods.
In total, we collect 150 annotations with 3 annotations per comparison. These results are shown in Table~\ref{tab_curiosities_results}.

\begin{table}[htbp]
\centering
\begin{tabular}{@{}lcccc@{}}
\toprule
                 & \textbf{Win \%} & \textbf{Relevance} & \textbf{Truthfulness} & \textbf{Fun Factor} \\ \midrule
Manually Curated & 33.33\%            & 0.71               & 1.78                  & 0.80                \\
Generated        & 66.67\%            & 1.56               & 1.50                  & 0.90                \\ \bottomrule
\end{tabular}%
\vspace{3mm}
\caption{Comparison between manually curated curiosities and curiosities generated by GPT-4.}
\label{tab_curiosities_results}
\end{table}

The results show that the annotators have a clear preference for the generated curiosities, despite their truthfulness value being lower than the manually curated ones. 
Regarding the fun factor, none of the methods exhibits a high score. This suggests that the concept of ``fun'' in this domain requires further exploration.
Nevertheless, the generated curiosities reduce the bottleneck of creating a manual set of curiosities and tend to be more fun and considerably more relevant.

\section{RQ3: Zero-Shot Conversation Ramblings}
\label{sec_zero_shot}
Often, users elaborate their responses or ask for side information, which conventional systems, with a strict dialogue flow, fail to respond.
TWIZ aims to improve the user experience by adapting \textit{on-the-fly to conversation ramblings} introduced by the user, through zero-shot approaches.

\paragraph{Zero-shot DST as Reading Comprehension}
\label{sub_zero_shot_dst}
A rapidly evolving dialogue system such as TWIZ requires DST modules that easily support the inclusion of new slots and intents, even with little to no available data. 
To achieve this, we follow \textit{Namazifar et al.}~\cite{qanlu} to cast the typical zero-shot DST task as a reading comprehension one. 
To train our models, we require QA examples. These are automatically extracted from unlabeled TWIZ data, in a self-supervised setting~\cite{diogot_sigir}.
Furthermore, during inference time, we require one question per slot and intent. These are created by prompting an LLM (\textit{text-davinci-003)} to generate one question per slot. This contrasts with typical approaches, where questions are either derived from templates or manually written~\cite{qanlu}.
We find that the strategy of pre-training the model using in-domain dialogues and LLM-based questions significantly improves performance, as detailed in our work~\cite{diogot_sigir}.

This model is used when new slots and intents are required to support new features, and while the team is gathering relevant, in-domain, data. When enough data is available, we pivot towards full-shot approaches (Section~\ref{sub_intent_detection}).

\paragraph{Zero-Shot Responses with TWIZ-LLM}
Generative Large Language Models trained with Human Feedback~\cite{instructgpt} possess a great ability to answer to a wide range of user ramblings.
After the release of ChatGPT and its open-source competitors~\cite{vicuna2023, alpaca}, it became clear that LLMs wrapped in strong guard rails~\cite{instructgpt, anthropic_rlhf} can respond to chit-chat, question-answering, and fallback intents in an appropriate and pleasant way.
In this context, we use the responses given by TWIZ-LLM (Section~\ref{sub_twiz_llm}) to respond to many of these requests.

\section{Trustworthy TaskBot Generative Vision and Language}
\label{sec_generative_vision_and_language}
Given the recent advancements in using generative methods for both text~\cite{vicuna2023, instructgpt, alpaca} and image generation~\cite{stable-diffusion}, a recent and very important research direction is how to guarantee trustworthy and consistent generations~\cite{q2, nl2vi}.
In our work, we make use of various generative methods and we always ensure that they are as trustworthy as possible through the implementation of several guardrails and verification methods.

In text generation tasks that are grounded in some input constraint, such as the \textit{Creative Cooking} that uses only valid tags for generating new recipes, the \textit{Task Promoter}, and the \textit{Curiosities} which receive a grounding Task/Wikipedia passage, are methods that can be further checked by a verification method such as \textit{True}~\cite{true} and \textit{Q2}~\cite{q2},
which use question generation/answering methods to ensure that the output is factually consistent with the input.

In the case of images, we take specific care in generating images that are \textit{Consistent} with the prompt through the use of a novel verification pipeline NL2VI~\cite{nl2vi} and further expand this work to generate images that are \textit{Coherent} along a sequence of steps.
These methods are based on the creation of a \textit{visual prompt} which is more suitable for image generation, followed by a T2I pipeline where images are verified through the use of VQA algorithms.

We believe that these methods are a step forward in diminishing the problems found in generative methods with the aim of providing more relevant and accurate information to the user.

\section{User Interaction Analysis}
\paragraph{User Behavior}
\label{sub_user_behavior}
Understanding user's interactions allows us to discover possible interaction bottlenecks or new conversational paths.

Looking at device type, we see 
that on average, ratings from headless interactions are higher than multimodal ones 3.47 vs. 3.27, respectively.
The use of multimodal devices also led to the increase of users using the on-screen buttons, rather than their voice, with over 35\% of the interaction turns being taps on the screen. 
Another typical user behavior is asking for commands that our bot cannot comply with, such as playing YouTube videos or music, which happens in 8.4\% of interactions. 
We also see that non-rated conversations tend to be very short, with an average of 2.35 vs. 7.46 turns for rated ones.  
When focusing on ratings and task execution, we see that 32\% of users who rate, start a task.
From these, 42.8\% start a recipe while the remaining 57.2\% do a WikiHow task. Additionally, they also rate significantly higher than those who don't start (3.86 vs. 3.30, respectively). Looking at the task type, users who start a recipe give an average rating of 4.07 whereas users who start a WikiHow have an average rating of 3.70.

\begin{figure}[t]
\centering\includegraphics[width=0.93\textwidth]{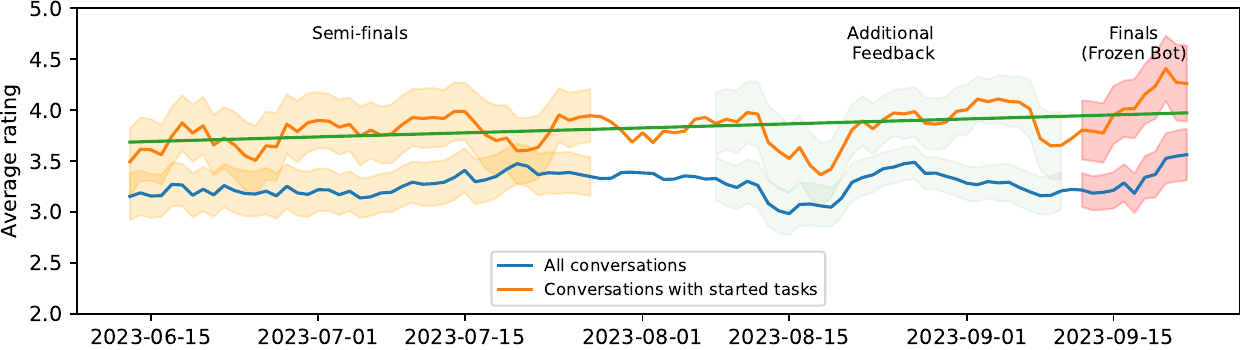}
    \caption{TWIZ's 7-Day average rating since the semi-finals for conversations with at least 3 turns.}
    \label{fig_ratings}
\end{figure}

\paragraph{Ratings Progression}
\label{rating_progression}
In Figure~\ref{fig_ratings}, we present the average 7-Day rating since the semi-finals. There are two major observations: 1) \textit{an increase in the user ratings throughout this period}, and 2) \textit{TWIZ's high effectiveness in conversations in which users start a task}. 
Moreover, a positive trend in the ratings can be observed from the start of the semi-finals. In some periods, ratings variability is high, specially in the middle of each stage's period, which we primarily attribute to intense system modifications. 

\begin{figure}[t]
    \centering
    \includegraphics[width=0.90\textwidth]{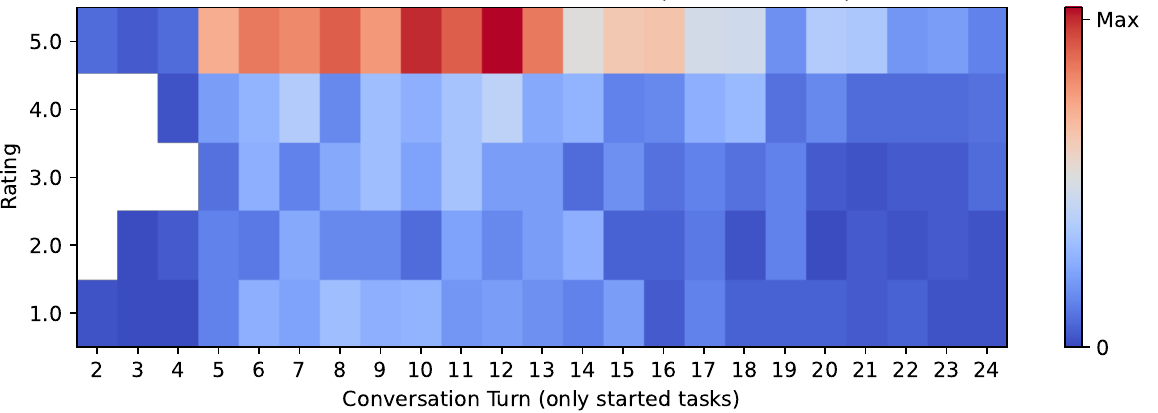}
    \caption{Conversations' ratings per turn when a task is started.
    }
    \label{fig_ratings2}
\end{figure}

\paragraph{Ratings per Conversation Length}
Finally, Figure~\ref{fig_ratings2}, depicts conversations rating per turn and provides further evidence of TWIZ bot's effectiveness in delivering the Alexa Prize TaskBot Challenge's main objective - guiding an user through a complex task. Namely, it can be seen that a large portion of conversations in which users start a task, get the maximum rating (top line).
These results highlight TWIZ bot's consistency and quality during a comprehensive time frame and for users who are using the bot in its full capacity.

\section{Conclusions and Future Work}
\label{sec_conclusions}
In this paper, we summarized TWIZ’s work on the second edition of the Alexa TaskBot Challenge. 
We built upon the strong foundation of the previous year and provided several new contributions:
\begin{itemize}
\item We focused on ensuring an engaging interaction by supporting \textit{natural and knowledgeable} dialogues.
To achieve this, we introduced the ``Creative Cooking'' feature, empowering users to craft their own custom recipes. Adding to this, we promote tasks in a positive way to increase user engagement, and provide navigational and contextual responses to user requests using TWIZ-LLM, an LLM trained specifically for the TaskBot domain.

\item We provided various \textit{multimodal stimuli} to the user. We expanded the curiosities paradigm and focused on improving multimodal interactions by generating task illustrations. In this last point, we proposed new methods to illustrate the task as well as its steps, with the added care to make illustrations consistent throughout the task. Still, in the multimodal stimuli, we also developed a video moment retrieval pipeline allowing video navigation by voice.

\item Finally, we created a \textit{robust} system to allow for more \textit{flexible} user interactions and analyzed the users' interactions with the system resulting in several relevant insights.
\end{itemize}

In future work, we aim to continue exploring the use of LLMs and expanding their utilization to all stages of the dialogue, as well as introducing new multimodal stimuli, while assessing their impact on the conversation.

\bibliographystyle{ACM-Reference-Format}
\bibliography{sample-base}

\clearpage
\appendix

\section{Creative Cooking: \textit{What’s in your fridge?} - Interaction}

\begin{figure}[htbp]
     \centering
     \begin{subfigure}[b]{0.45\textwidth}
         \centering
         \includegraphics[width=\textwidth]{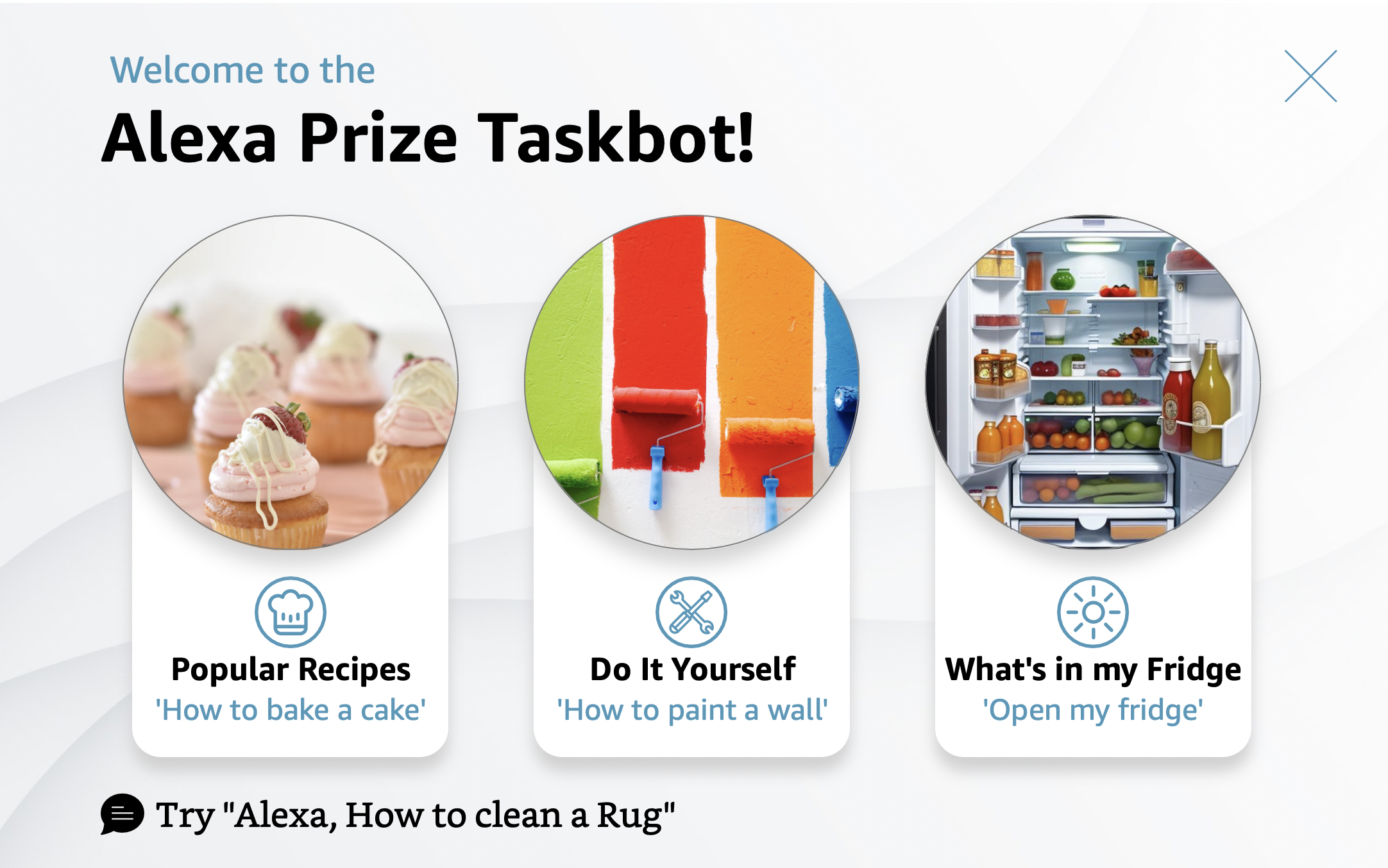}
    \vspace{3mm}
     \end{subfigure}
     \quad \quad
     \begin{subfigure}[b]{0.45\textwidth}
         \centering
         \includegraphics[width=\textwidth]{figures/creative_cooking/add_ingredients.png}
         \vspace{3mm}
     \end{subfigure}
     \begin{subfigure}[b]{0.45\textwidth}
         \centering
         \includegraphics[width=\textwidth]{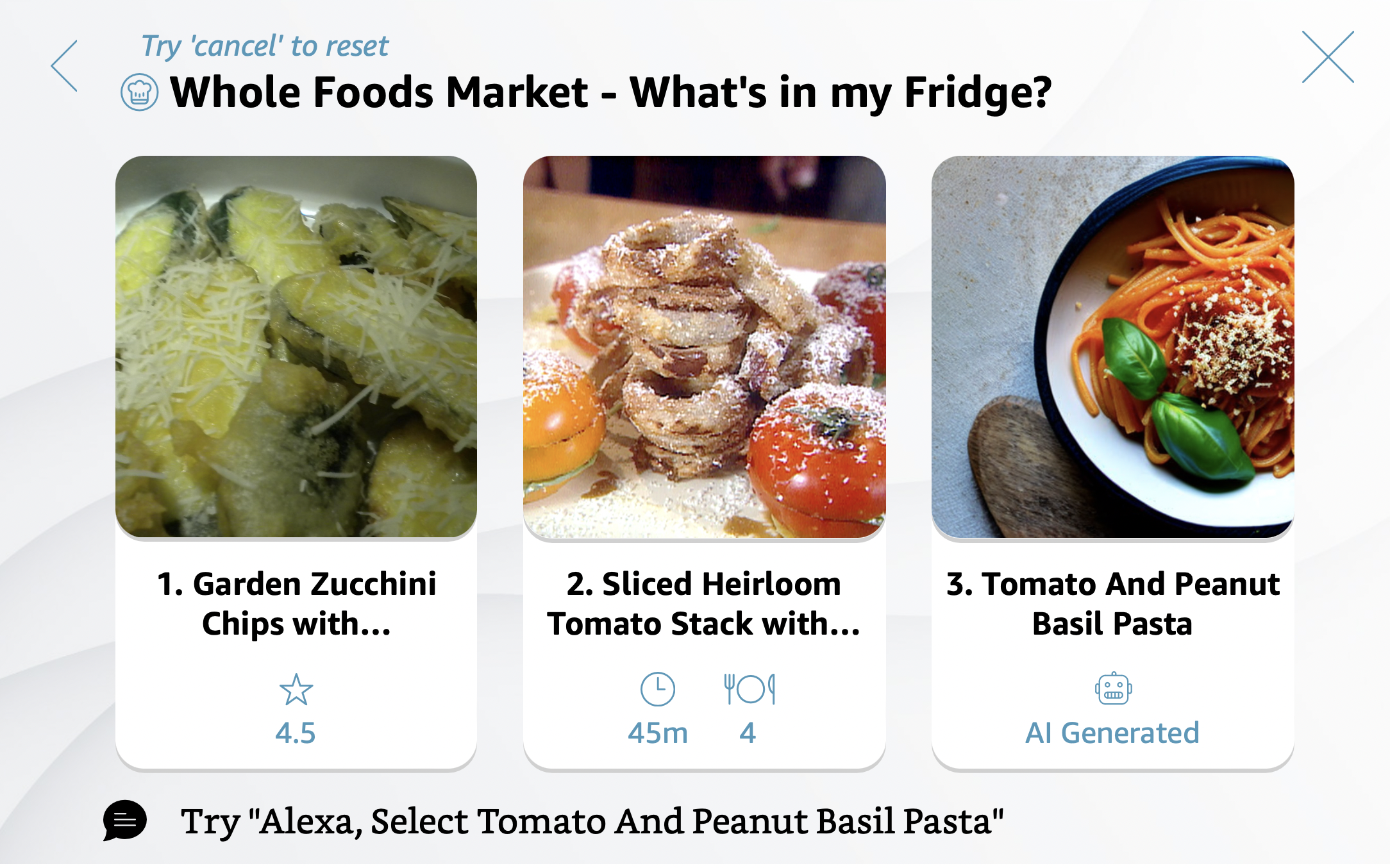}
     \end{subfigure}
     \quad \quad
     \begin{subfigure}[b]{0.45\textwidth}
         \centering
         \includegraphics[width=\textwidth]{figures/creative_cooking/generated_task_overview.png}
     \end{subfigure}
     \caption{Creative cooking interaction examples.}
     \label{fig_creative_cooking}
\end{figure}

\section{Task Promoter}
\label{app_repro_training}

\subsection{Example Outputs}

\begin{table}[htbp]
\centering
\begin{tabular}{@{}p{0.17\textwidth}p{0.3\textwidth}p{0.5\textwidth}@{}}
\toprule
\textbf{Recipe Title} & \multicolumn{1}{c}{\textbf{Ingredients}} & \multicolumn{1}{c}{\textbf{Promoter Output}} \\ \midrule
Lemony Lentil Salad &
  [cauliflower, green lentils, couscous, lemon, garlic, honey, extra virgin olive oil, parsley, feta cheese] &
  Lemony Lentil Salad: A zesty, plant-based meal that's packed with flavor and nutrients. Cauliflower, lentils, and couscous come together in a tangy dressing for a satisfying and healthy salad. Top with feta for a pop of flavor and crunch. \\ \midrule
Garlic and Cheese Popovers &
  [butter, chives, lemon, egg, garlic, milk, flour, salt, ground pepper, parsley] &
  Upgrade your dinner game with these garlic and cheese popovers. Perfect for a cozy night in or a special occasion, they're sure to impress. Trust us, your taste buds will thank you! \\ \midrule
Apple Crumble &
  [apple, sugar, lemon, ground cinnamon, nutmeg, ginger root, salt, flour, oats, dark brown sugar, ground cinnamon, butter] &
  This is my favorite homemade apple crumble recipe because it's simple to make but comes out perfectly fluffy and light! \\ \bottomrule
\end{tabular}%
\vspace{2mm}
\caption{Promoter example outputs.}
\label{tab_promoter_examples}
\end{table}

\subsection{Training Details}
\label{app_promoter_training_details}
To train the \textit{RePro} model, we used 5k recipe descriptions along with their ingredients and steps. For the model, we used a GPT-2-774M~\cite{radford_gpt2}, trained on instruction data~\cite{DatabricksBlog2023DollyV2}. 
We then further fine-tuned this model with Direct Preference Optimization (DPO)~\cite{rafailov2023direct}. Given that DPO requires preference data, we prompted an LLM~\cite{vicuna2023} for recipe descriptions and discarded any that were hallucinated (e.g. wrong number of ingredients mentioned, or wrong ingredients mentioned).

\section{TWIZ-LLM}
\label{app_twiz_llm}

\subsection{Dataset}
\label{app_twiz_llm_dataset_creation}

\begin{table}[htbp]
\centering
\resizebox{0.94\textwidth}{!}{%
\begin{tabular}{p{0.07\textwidth}p{0.27\textwidth}p{0.55\textwidth}}
\toprule
\multicolumn{3}{l}{\textbf{Task Title:} Creamy Pesto Chicken Soup with Avocado}                                  \\ \midrule
\textbf{Turn \#} & \textbf{User}                  & \textbf{System}                                                  \\ \midrule
1                & -                              & Step 1: Pour the Chicken Stock in a pan and add Garlic. \\
2 & Got it done.        & Step 2: Bring the stock to a boil over high heat and then add the Chicken Breast. \\
3                & I do not have chicken breasts. & No problem, you can also use Turkey.                             \\
$\ldots$ & $\ldots$ & $\ldots$                                                                                         \\
7 & Tell me a fun fact. & In the US, on the 31st of July, it's national avocado day.              \\
8 & Next.               & Step 8: Scoop the avocado chicken soup into bowls. \\ 
9                & Next step.                     & We have reached the end of the task.                             \\
10               & Stop.                          & Glad I could help you! See you again soon!                       \\ \bottomrule
\end{tabular}%
}
\vspace{2mm}
\caption{Example conversation from TWIZ LLM dataset with manually-crafted utterances.}
\label{tab_example_conversation}
\end{table}

\textbf{Questions.} For the questions, we prompted OpenAI's API\footnote{specifically, \textit{text-davinci-003}} to generate both question and answer, given the step text.  

\textit{\textbf{Ingredients Replacement}} - To simulate user utterances requesting an ingredient replacement, we apply a set of templates and fill in with an ingredient of a step. The system answer is also template-based, by accessing an external database of ingredient substitutes\footnote{\url{https://foodsubs.com/}}.

\textit{\textbf{Definition Question}} - We extract noun phrases from the current step using spaCy and randomly pick one to fill in a ``What is?'' type question. The answer is extracted from a dictionary\footnote{\url{https://github.com/wordset/wordset-dictionary}}, using the intersection between the step's text and the various definitions to select the best one.

\subsection{Example Prompts} 
\label{app_example_prompts}

Tables~\ref{tab:prompt_example} and \ref{tab:vicuna_prompt_example} show the prompts used to train the OPT and Vicuna models, respectively.

\begin{table}[htbp]
\centering
\resizebox{0.95\textwidth}{!}{%
\texttt{%
\footnotesize
\begin{tabular}{p{\linewidth}} \toprule
Below is a recipe and the steps to make it, the step of the recipe that the user is doing, and part of the dialogue between you and a user about this recipe. Write a response that appropriately responds to the user request. Respond with a \textit{neutral} tone. \\ \\ 
Recipe: \\ 
Garlicky Broccoli. Step 1: Wash Broccoli in water and shake off excess water. Chop top (head) of the Broccoli into bite-sized pieces. Step 2: Heat Extra-Virgin Olive Oil in a large nonstick skillet over medium-high heat. Step 3: Add Garlic to the pan by crushing with a garlic press. 
\textit{(other steps omitted)}\\
\\ 
Current Step: \\ 
Step 1: Wash Broccoli in water and shake off excess water. Chop top (head) of the Broccoli into bite-sized pieces. \\ \\ 
Dialog: \\ 
User: Is it necessary to wash the broccoli before chopping? System: Yes, it is important to wash the broccoli before chopping to remove any dirt or debris. User: next. \\ \\
Response: \\
Heat Extra-Virgin Olive Oil in a large nonstick skillet over medium-high heat. \\ 
\bottomrule
\end{tabular}
}
}
\vspace{2mm}
\caption{Example of the input prompt to the OPT model.\label{tab:prompt_example}}
\end{table}

\begin{table}[htbp]
\centering
\resizebox{0.95\textwidth}{!}{%
\texttt{%
\footnotesize
\begin{tabular}{p{\linewidth}} \toprule
<|prompter|> I will give you a recipe and I want you to help me do it step by step. Please use a \textit{neutral} tone of voice. Recipe: Garlicky Broccoli. Step 1: Wash Broccoli in water and shake off excess water. Chop top (head) of the Broccoli into bite-sized pieces. Step 2: Heat Extra-Virgin Olive Oil in a large nonstick skillet over medium-high heat. Step 3: Add Garlic to the pan by crushing with a garlic press. 
\textit{(other steps omitted)} This is the current step: Wash Broccoli in water and shake off excess water. Chop top (head) of the Broccoli into bite-sized pieces. <|endoftext|> <|assistant|> ok! <|endoftext|> <|prompter|> Is it necessary to wash the broccoli before chopping? <|endoftext|> <|assistant|> Yes, it is important to wash the broccoli before chopping to remove any dirt or debris. <|endoftext|> <|prompter|> next. <|endoftext|> <|assistant|> Heat Extra-Virgin Olive Oil in a large nonstick skillet over medium-high heat. <|endoftext|> </s> \\
\bottomrule
\end{tabular}
}
}
\vspace{2mm}
\caption{Example of the input prompt to the Vicuna model.\label{tab:vicuna_prompt_example}}
\end{table}

\section{Automated Testing} 
\label{app_automated_testing}

\subsection{User Simulator} 
\label{sec_user_simulator}

As TWIZ becomes increasingly more complex and undergoes constant changes, ensuring its correct behavior during testing becomes more challenging. To address this bottleneck in development, we created a testing tool that utilizes CoBot's interactive mode~\cite{cobot_2018} to simulate user interactions. This user simulation can be conducted using either previously recorded user interactions or generated ones. By making the testing process automatic, the testing time can be shortened, which accelerates the R\&D cycle.

While previously recorded user interactions help check if previous features weren't compromised while adding new ones, the generated user interactions allow to explore previously unseen conversation flows, allowing for the creation of a more robust assistant. 
In order to generate these interactions, our user simulation tool leverages the dialogue manager state-machine pattern implementation (Section \ref{sub_dialog_manager}) and user conversation transcripts. The transcripts provide the necessary data to compute state transition probabilities and enable the creation of a user utterance bank linked to specific state and event pairs. 

\begin{figure}[ht]
    \centering
    \includegraphics[width=0.6\textwidth]{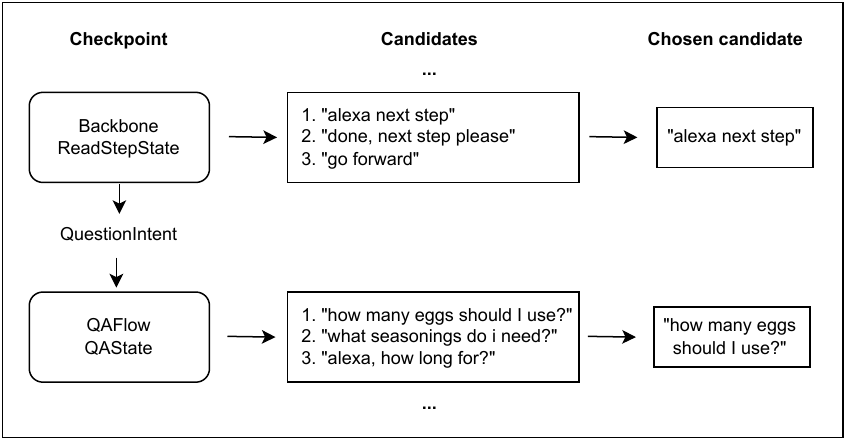}
    \caption{User simulation procedure.}
    \label{fig:candidate-picking}
\end{figure}

The user simulation generation process starts with the tool initiating an interaction with TWIZ. For each turn of the conversation, an event is selected based on the computed transition probabilities and the current state of the dialogue. Then, the user utterances related to the current state and sampled event are ranked by similarity using sentence embeddings~\cite{SentenceBert}. 
This ranking is performed by calculating the contextual embeddings of the ongoing conversation and determining the cosine similarity with the contextual embeddings of all potential utterances. This makes it so that the most relevant and contextually appropriate utterance is chosen, resulting in compelling and cohesive dialogues. 
An example is provided in Figure~\ref{fig:candidate-picking}.

\subsection{Rating Prediction for Conversational Task Assistants}
Due to the complex interactions between the user and the system, errors are prone to happen which in turn lead to user dissatisfaction and low ratings. Being able to predict the rating of an interaction is thus a critical step to understand the system's shortcomings, and act accordingly~\cite{amazon_2023_rating_prediction, choi_rating_prediction}.
Emphasizing this, on average less than 5\% of conversations have an associated rating, making it difficult to decide which conversations should be prioritized for analysis. 
Moreover, rating prediction complements the creation of the simulated interactions (Section~\ref{sec_user_simulator}), allowing to automatically rate these generated interactions.

Inspired by the work from \textit{Choi et al.}~\cite{choi_rating_prediction} in SocialBot, we developed a rating prediction model specific to the TaskBot setting~\cite{rafael_rating_prediction}, where the aim is to predict a rating given an entire conversation. 
Our model is based on the Transformer~\cite{vaswani2017attention} architecture and combines both textual features, i.e., the conversation between the user and system, and user-behavior features, such as the number of fallbacks/steps read, and other general and Taskbot-specific features.
To evaluate our model, we used conversations and ratings collected during the first edition of the Alexa Prize TaskBot challenge. 
In this setting, the model achieves an accuracy of 70\% in a binary rating classification task. 
The results demonstrated the utility of automatic rating prediction, revealing insights such as the significance of starting a task and the increased importance of the latter part of the conversation~\cite{rafael_rating_prediction}.

\end{document}